\documentclass[10pt,a4paper,twocolumn]{article}

\usepackage[top=0.72in,bottom=0.72in,left=0.70in,right=0.70in]{geometry}
\usepackage[T1]{fontenc}
\usepackage[utf8]{inputenc}
\usepackage{lmodern,amsmath,amssymb,booktabs,tabularx,array}
\usepackage{microtype}
\usepackage{xcolor}
\usepackage{tikz,pgfplots}
\usepackage[font=small,labelfont=bf,justification=raggedright,singlelinecheck=false]{caption}
\usepackage[hidelinks]{hyperref}

\usetikzlibrary{arrows.meta,positioning}
\pgfplotsset{compat=1.18}

\definecolor{evidence}{HTML}{176B73}
\definecolor{rehearsal}{HTML}{A66325}
\definecolor{ink}{HTML}{23333B}
\definecolor{softgray}{HTML}{A9B4B9}

\newcommand{\tracepanel}[3]{%
  \begingroup\setlength{\fboxsep}{6pt}%
  \noindent\fcolorbox{#1!55}{#1!4}{%
    \begin{minipage}[t]{\dimexpr\linewidth-2\fboxsep-2\fboxrule\relax}
    \footnotesize\raggedright\textbf{#2}\par\vspace{3pt}#3
    \end{minipage}}\endgroup}

\pgfplotsset{paperaxis/.style={
  tick label style={font=\scriptsize},
  label style={font=\small},
  title style={font=\small\bfseries},
  axis line style={ink!45},
  tick style={ink!40},
  grid style={black!8},
  legend style={font=\scriptsize,draw=none,fill=none},
  axis x line=bottom,
  axis y line=left}}

\title{MamaReason: Unfinished Reasoning and Answer Recovery\\in Clinical Language Models}

\author{%
\small
\begin{tabular}{c}
Thanni Adewuyi$^{1}$ \quad
Adepoju Jeremiah Moyondafoluwa$^{1}$ \quad
Dorcas Amapu$^{1}$ \quad
Dieko Whenu$^{1}$ \\[0.35em]
Sylvia Essessien$^{1}$ \quad
Oko-Oba Sandra$^{1}$ \quad
Pelumi Onafadeji$^{1}$ \quad
Lily Aimas-Love$^{1}$ \\[0.35em] 
Samuel Omolaiye$^{1}$ \quad
AbdulQahhar Okenla$^{1}$ \quad
Abiodun Adereni$^{1}$ \\[0.5em]
\footnotesize
$^{1}$HelpMum Africa
\end{tabular}
}
\date{}

\begin{document}
\maketitle
\begin{abstract}
Answer continuation often returns a diagnosis already stated in unfinished clinical reasoning. We study this behavior in MamaReason, adapted from dense Gemma 4 31B, on MamaBench's 217 paired maternal and child-health presentations. An explicit-commitment extractor identifies a diagnosis in 75 of 94 failed RL first passes and 75 of 93 failed SFT first passes. The continuation matches that diagnosis in 67/75 RL cases and 58/75 SFT cases, using diagnosis strings alone. Reading these commitments without additional generation recovers 30 of the 40 RL pairs restored by continuation under the archived correctness labels. Recurrence controls distinguish routine mentions from repeated answer selection. An offline second-commitment trigger measures the tradeoff between earlier readout and paired accuracy. The resulting decomposition separates diagnostic content already present in a trace, completion of an accepted answer, and paired diagnostic correctness. For clinical language-model evaluation, these are distinct outcomes that become visible when the same reasoning trace is examined through multiple readouts.
\end{abstract}
\section{Introduction}
A clinical language model can settle on a diagnosis in its reasoning yet fail to return the answer expected by an evaluation harness. When the harness then asks the model to finish, a higher score can reflect several changes: exposing a diagnosis already present, revising that diagnosis, or supplying a missing clinical qualifier. These possibilities matter because clinical reasoning benchmarks usually score the final answer, while the distinction between diagnostic content and answer completion remains hidden.

Medical QA benchmarks assess clinical knowledge and biomedical evidence use \cite{medqa,pubmedqa,medmcqa}; broader evaluations examine open-ended responses and diagnostic assistance \cite{multimedqa,healthbench,medhelm,ddx,amie}. Paired evaluations add a complementary requirement: success on related presentations whose discriminating details support different diagnoses \cite{contrast,checklist,mamabench,reasoningknowledge}. We use MamaBench's maternal and child-health pairs to examine how diagnoses already stated in a reasoning trace relate to the answers and scores produced by subsequent readout.

Our central finding is label-independent. In MamaReason, adapted from dense Gemma 4 31B through SFT and clinical-content RL \cite{gemma4}, a frozen rule set extracts explicit commitments from 75/94 failed RL first passes and 75/93 failed SFT first passes. Continuation returns the same normalized diagnosis in 67/75 and 58/75 extracted cases, respectively. Across all failures, this gives observed matches in 67/94 RL cases (71.3\%) and 58/93 SFT cases (62.4\%), with abstentions retained in the denominator. Most extractable failures therefore contain the diagnosis that the continuation eventually supplies.

We compare three readouts of the saved first passes: the existing JSON parser, an extraction-only last-diagnosis fallback, and the actual conditional continuation. Under the archived correctness labels, the RL fallback recovers 30 of the 40 pairs restored by continuation; 29 of these recoveries use matching continuation diagnoses and their saved verdicts. Figure~\ref{fig:emission_lead} connects this scored recovery to the diagnosis-preservation result. Recurrence controls and prefix readouts then characterize the traces surrounding these commitments. This extends the study of test-time reasoning and budget forcing \cite{s1,tokenbudget,budgetthinker} to the clinical distinction between reaching a diagnosis and delivering an accepted answer.

\begin{figure*}[t]
\centering
\begin{minipage}[t]{.48\textwidth}
\centering
\begin{tikzpicture}
\begin{axis}[paperaxis,width=\linewidth,height=5.0cm,ybar,bar shift=0pt,bar width=20pt,
 ymin=0,ymax=85,xmin=-.55,xmax=3.95,ylabel={RL robust accuracy (\%)},
 xtick={0,1.4,2.4,3.4},xticklabels={Direct,Parser,Commitment,Continuation},
 x tick label style={font=\scriptsize,rotate=15,anchor=east},ymajorgrids=true]
\addplot[fill=softgray,draw=none,bar shift=0pt] coordinates {(0,53.5)};
\addplot[fill=rehearsal!75,draw=none,bar shift=0pt] coordinates {(1.4,45.2)};
\addplot[fill=ink!75,draw=none,bar shift=0pt] coordinates {(2.4,59.447)};
\addplot[fill=evidence,draw=none,bar shift=0pt] coordinates {(3.4,63.6)};

\draw[ink!55,densely dashed] (axis cs:.65,0)--(axis cs:.65,80);
\node[font=\tiny,align=center] at (axis cs:0,77) {Separate\\generation};
\node[font=\scriptsize] at (axis cs:2.4,77) {Shared first pass};
\node[font=\scriptsize,anchor=south] at (axis cs:0,53.5) {53.5};
\node[font=\scriptsize,anchor=south] at (axis cs:1.4,45.2) {45.2};
\node[font=\scriptsize,anchor=south] at (axis cs:2.4,61.2) {59.4};
\node[font=\scriptsize,anchor=south] at (axis cs:3.4,63.6) {63.6};
\end{axis}
\end{tikzpicture}
\end{minipage}\hfill
\begin{minipage}[t]{.48\textwidth}
\centering
\tracepanel{ink}{Diagnosis preservation before continuation}{%
\textbf{RL:} commitments extracted from 75/94 failures; 67/75 match the continuation.\par\smallskip
\textbf{SFT:} commitments extracted from 75/93 failures; 58/75 match the continuation.\par\smallskip
These matches use diagnosis strings alone.\par\smallskip
\textbf{Scored recovery:} the RL fallback recovers 30/40 continuation-restored pairs under archived verdicts.}
\end{minipage}
\caption{Diagnosis preservation and paired recovery. The direct-answer reference is generated separately; the other bars share the same bounded-thinking first pass. Continuation adds a conditional answer-generation step. Commitment uses the completed diagnostic adjudication. All robust scores use 217 pairs.}
\label{fig:emission_lead}
\end{figure*}
The paper contributes:
\begin{itemize}
\setlength{\itemsep}{2pt}
\item A label-independent measurement of agreement between explicit first-pass commitments and continuation diagnoses, with extraction coverage reported for both checkpoints.
\item A paired-correctness decomposition across parser, commitment, and continuation readouts, including uncertainty for comparisons with direct answering.
\item Recurrence controls and interval answer-availability measurements that characterize repeated answer selection and late completion.
\end{itemize}
\section{Related Work}
\subsection{Medical QA and diagnostic evaluation}
MedQA and MedMCQA evaluate medical examination knowledge, while PubMedQA tests interpretation of biomedical abstracts \cite{medqa,medmcqa,pubmedqa}. Thapa et al.\ separate knowledge-heavy from reasoning-heavy biomedical questions and examine misleading initial hypotheses \cite{reasoningknowledge}. Broader clinical evaluations include long-form answer quality in MultiMedQA, physician-authored rubrics in HealthBench, a taxonomy of medical tasks in MedHELM, and diagnostic assistance in differential-diagnosis studies and AMIE \cite{multimedqa,healthbench,medhelm,ddx,amie}. MamaBench supplies a paired maternal and child-health diagnosis setting \cite{mamabench}. We connect its final-answer scores to diagnostic content in the preceding trace.

\subsection{Counterfactual robustness and behavioral testing}
Contrast sets probe local decision boundaries through meaningful test-instance changes \cite{contrast}. Counterfactually augmented data uses coherent edits to expose reliance on patterns that fail under revision \cite{counterfact}, and CheckList organizes tests around capabilities and expected behavior \cite{checklist}. MamaBench applies this perspective to related clinical presentations. Its pairs can change several observations, making joint correctness a test of sensitivity to the benchmark's evidence changes.

\subsection{Synthetic supervision and medical adaptation}
Self-Instruct combines generated instructions with filtering \cite{selfinstruct}; STaR iterates rationale generation and training on correct-answer trajectories \cite{star}. In medicine, HuatuoGPT-o1 uses verifiable problems for reasoning supervision and RL, while MedReason grounds reasoning paths in medical knowledge graphs \cite{huatuo,medreason}. MamaReason uses separate construction routes for structurally cleaned SFT demonstrations and reference-grounded RL vignettes. The distinction between generating, filtering, and verifying supervision is relevant to both routes.

\subsection{Reasoning traces and overthinking}
Chain-of-thought prompting elicits intermediate steps, and self-consistency aggregates answers across sampled paths \cite{cot,selfconsistency}. Studies of intrinsic self-correction and overthinking examine how later reasoning can preserve, improve, or degrade an earlier solution \cite{selfcorrection,overthinking}. Chen et al.\ assess the efficiency of successive solution rounds \cite{overthinking}; our clinical analysis examines whether a continuation preserves an earlier diagnosis or changes a consequential qualifier.

Visible reasoning also raises questions of faithfulness. Turpin et al.\ and Chen et al.\ show that generated explanations can omit influences on predictions \cite{faithfulness,reasoningfaithfulness}. Lanham et al.\ test reasoning through interventions, including early answering after truncation \cite{measuringfaithfulness}. Their active intervention asks what answer a model produces from an intermediate prefix. Our passive readout asks what diagnosis is already explicitly stated in the saved text. This distinction locates the present study between analysis of intermediate reasoning and inference-time answer control.

\subsection{Inference control and structured answers}
Effective allocation of test-time computation depends on problem difficulty \cite{testcompute}. s1 controls deliberation through budget forcing \cite{s1}; Token-Budget-Aware LLM Reasoning and BudgetThinker adapt reasoning to available computation \cite{tokenbudget,budgetthinker}. PICARD and guided generation enforce formal output structure during decoding \cite{picard,guided}. These methods act on generation. By holding the first-pass text fixed, our comparison separates existing diagnostic assertions from the answer supplied by a continuation, providing a reference point for active early-answering and stopping strategies.

\subsection{Reward optimization and verification}
Instruction tuning and feedback-based RL adapt language-model behavior \cite{instructgpt}. PPO uses clipped policy updates, and GRPO uses group-relative rewards in place of a learned critic \cite{ppo,deepseekmath}. Process supervision evaluates intermediate steps, while outcome supervision evaluates completed trajectories \cite{verify}; Dr.\ GRPO and DAPO study further optimization choices \cite{drgrpo,dapo}. MamaReason's clinical reward covers diagnosis, findings, urgency, and management. Benchmark diagnosis scoring and clinician review provide separate assessments, each with its own criteria. Work on LLM-based evaluation motivates examining the consistency of saved diagnostic verdicts \cite{llmjudge}.

\section{Methodology}\label{sec:methods}
\subsection{Model and study setting}
We introduce MamaReason as the clinical model lineage for this evaluation study, adapted from the dense Gemma 4 31B model \cite{gemma4}. We examine its SFT checkpoint and the checkpoint produced by one subsequent clinical-content RL run. The analysis combines saved generations with archived and additional diagnostic verdicts. We report the data construction, inference protocols, and readout rules following the emphasis on transparent clinical-LLM reporting in TRIPOD-LLM \cite{tripodllm}.

\subsection{Training context}
The SFT checkpoint uses clinical reasoning demonstrations with separate thought and visible-answer channels: 10,122 training examples and 533 validation examples. RL starts from that checkpoint and uses 1,000 vignettes spanning 111 conditions. MamaBench supplies the evaluation cases. Detailed preparation, sampling, and cleaning are in Appendix~\ref{app:training} and Table~\ref{tab:data}.

\subsection{Adaptation summary}
Both stages use quantized LoRA adaptation \cite{qlora,lora}. RL starts from the SFT checkpoint, samples four completions per prompt, and combines rule-based diagnosis, finding, urgency, and management rewards. Finding and management coverage can use thought text; the objective has no explicit repetition or length penalty. Appendix~\ref{app:training} gives the reward equations and optimization settings.

\subsection{Paired evaluation and answer emission}
MamaBench contains 217 pairs, or 434 clinical narratives, processed independently \cite{mamabench}. The principal prompt covers obstetric, gynecologic, neonatal, and pediatric diagnosis and allows non-obstetric answers. The main comparison uses this prompt in both checkpoints.

Table~\ref{tab:protocols} gives the output caps and answer procedures. Decoding is greedy, with temperature 0 and top-$p=1$. Bounded thinking triggers a continuation after a JSON parsing failure, blank prediction, or template answer. The continuation closes the thought channel if necessary and requests schema-constrained JSON, with an unconstrained fallback. Accepted first-pass predictions are retained. Raw thought and continuation text are stored separately.

\begin{table}[!htbp]
\centering\small
\caption{Principal inference protocols and maximum output lengths.}
\label{tab:protocols}
\begin{tabularx}{\linewidth}{@{}p{0.24\linewidth}p{0.19\linewidth}X@{}}
\toprule
Protocol & Maximum output & Emission procedure\\
\midrule
Direct answer & 1,024 tokens & Thinking disabled through the chat template.\\
Bounded thinking & 3,000 + 512 tokens & Conditional thought closure and answer-rescue continuation.\\
\bottomrule
\end{tabularx}
\end{table}

A temperature-zero diagnostic evaluator receives the reference diagnosis and prediction. Its criteria accept synonymous names and require clinically consequential qualifiers. Blank or placeholder predictions and explicit parsing/evaluator errors count as incorrect. Record-level verdict consistency is documented in Appendix~\ref{app:provenance}.

Let $PP$, $PF$, $FP$, and $FF$ denote both correct, base-only correct, counterfactual-only correct, and neither correct. For $N$ pairs,
\begin{equation}
\operatorname{Acc}_{\mathrm{base}}=\frac{PP+PF}{N},\qquad
\operatorname{Acc}_{\mathrm{rob}}=\frac{PP}{N},
\end{equation}
\begin{equation}
\operatorname{BTR}=\frac{PF}{PP+PF}.
\end{equation}
Robust accuracy requires both presentations correct. Bias Trap Rate (BTR) is the counterfactual failure rate among base-correct pairs, defined when $PP+PF>0$. All principal comparisons contain $N=217$ complete pairs.

\subsection{Replaying the first-pass readout}
\label{sec:replay}
For rescued records, we remove the marked continuation and its synthetic channel delimiter, then apply the evaluator's JSON parser and blank/template gate to the first-pass text. A prediction is \emph{available} when the parser returns a nonempty, non-placeholder diagnosis. This includes JSON objects recoverable from an unclosed thought channel. For available first passes, we verify agreement with the saved prediction and retain its verdict; unavailable predictions score zero.

Continuation markers, replayed first-pass failures, and successful-rescue flags agree exactly in the principal runs: 94 RL and 93 SFT cases. Every marked continuation yields an accepted prediction. Comparing first-pass and final scores therefore holds the original first-pass generation fixed.

For pair $i$, let $a_i$ indicate that both first-pass predictions are available and $c_i$ indicate joint correctness. Pair availability and conditional robust accuracy are
\begin{equation}
\begin{aligned}
C_{\mathrm{pair}}&=\frac{\sum_i a_i}{N},\\
A_{\mathrm{cond}}&=\frac{\sum_i a_i c_i}{\sum_i a_i}.
\end{aligned}
\end{equation}
The unconditional score is
\begin{equation}
A_{\mathrm{first}}=C_{\mathrm{pair}}A_{\mathrm{cond}}.
\label{eq:coverage}
\end{equation}
All unconditional scores retain 217 pairs.

\subsection{Token reconstruction and prefix availability}
We re-tokenize saved text with a public base tokenizer, preserving special tokens and adding none. Decoding reproduces the text exactly for all 868 principal bounded-thinking first passes. Appendix~\ref{app:tokenizer} records reconstruction details. We describe a failure as \emph{cap-consistent} when its reconstructed length is within one token of 3,000, and detect a narrow exact-cycle signature when an 8--32-token block repeats at least three times consecutively. Thought-channel closure and explicit end-of-turn markers are recorded separately.

At prefix lengths of 250, 500, 750, 1,000, 1,500, 2,000, 2,500, 2,750, and 3,000 tokens, we reapply the same parser and gate. For case $k$, let $v_k(t)$ indicate availability at prefix $t$. The answer yield over an observed interval is
\begin{equation}
Y_{\ell,u}=\frac{\sum_k[1-v_k(\ell)]v_k(u)}{\sum_k[1-v_k(\ell)]}.
\end{equation}
We compute yields within each saved run; the longer historical configuration is described in Appendix~\ref{app:secondary}. Their denominators retain earlier-stopped outputs; we also report the subset whose saved length exceeds $\ell$. These are readouts of fixed text, so availability can precede the end of a saved generation.

\subsection{Commitment extraction and recurrence}
The extractor selects the last eligible diagnosis assertion or answer-selection statement. It scans first-pass text alone, using a frozen set of cues and exclusions specified in Appendix~\ref{app:extractor}. With no eligible assertion, it abstains. Rules were developed on RL traces and applied unchanged to the SFT failures.

We measure recurrence in the final 500 thought tokens. Diagnosis mentions are word-boundary matches after case and punctuation normalization; explicit assertions additionally require an eligible selection cue. Two matches define recurrence. The control uses the accepted diagnosis in 340 available RL first passes, excluding visible-answer text. For failed outputs, we examine both the extracted commitment in 75 eligible cases and the continuation diagnosis in all 94 cases. A complementary whole-trace repeated-four-gram statistic, $r_4$, is defined in Appendix~\ref{app:traces}.

\subsection{Offline second-commitment readout}
We apply the frozen assertion rules to prefixes of each observed thought channel. The trigger is the first prefix containing two eligible assertions of the same normalized diagnosis. It selects the diagnosis at the second assertion. If the thought channel closes first or the rule never triggers, the original first-pass parser result is retained. The detector uses only text available at that prefix.

We record trigger coverage, agreement with the saved final prediction, and paired correctness of this extraction-only policy. The text reduction is the fraction of reconstructed first-pass tokens beyond the selected prefixes, summed over all 434 cases. It excludes rescue-continuation tokens. This offline analysis reads the selected diagnosis directly; live continuation after a trigger is a separate experiment.

\subsection{Correctness scoring and paired uncertainty}
The extraction-only fallback retains accepted first-pass predictions and fills failed cases with extracted commitments. Matching diagnoses reuse the archived verdict for that case. The remaining 20 last-commitment candidates (five RL and 15 SFT) are scored with the same diagnostic evaluator, prompt, and temperature-zero settings as the original benchmark run. Three additional candidates from the early-readout experiment are scored identically. All extracted candidates now have verdicts; abstentions remain unavailable. Diagnosis-string agreement is measured separately from these correctness labels. Appendix~\ref{app:adjudication} records the added assessments.

We resample the same 217 pair identifiers 20,000 times with replacement, using seed 42, and report percentile 95\% bootstrap intervals \cite{bootstrap}. Exact two-sided McNemar tests use gains and losses in joint correctness. We apply both procedures to first-pass-minus-direct and fallback-minus-direct comparisons. Continuation recovery is monotone because only failed first passes receive a continuation. Tests are exploratory and unadjusted for multiple comparisons. The exact test conditions on discordant pairs, which can produce a different threshold decision from the percentile bootstrap near zero.

\section{Results}
\subsection{Continuations preserve earlier diagnostic commitments}
The extractor identifies commitments in 75/94 failed RL first passes (79.8\%) and 75/93 failed SFT first passes (80.6\%), abstaining in 19 and 18 cases. Continuation matches the extracted diagnosis in 67/75 RL cases (89.3\%) and 58/75 SFT cases (77.3\%). Counting matches over all failures gives 67/94 (71.3\%) and 58/93 (62.4\%). These results use normalized diagnosis strings alone and show that the continuation commonly returns a diagnosis already stated in the unfinished trace.

\subsection{Answer availability changes the protocol comparison}
\label{sec:emission_results}
Table~\ref{tab:emission} reports availability and correctness for all principal conditions. The RL first pass is jointly correct on 98/217 pairs (45.2\%), compared with 116/217 (53.5\%) for direct answering: $-8.3$ points (paired-bootstrap 95\% interval, $-14.3$ to $-2.3$; 13 gains, 31 losses; McNemar $p=0.0096$). For SFT, the difference is also $-8.3$ points (43.8\% versus 52.1\%; interval, $-14.3$ to $-2.8$; 12 gains, 30 losses; $p=0.0079$).

\begin{table*}[t]
\centering\small
\caption{Answer availability and correctness over 434 cases and 217 pairs. First-pass and end-to-end readouts share the bounded-thinking first pass; the latter adds conditional continuation. Direct answering is generated separately. An available pair has two accepted predictions.}
\label{tab:emission}
\setlength{\tabcolsep}{4pt}
\begin{tabular}{@{}llrrrrr@{}}
\toprule
Checkpoint & Readout & Available cases & Available pairs & Correct cases & Correct pairs & Robust (\%)\\
\midrule
SFT & Direct answer & 434 & 217 & 307 & 113 & 52.1\\
SFT & Thinking first pass & 341 & 133 & 279 & 95 & 43.8\\
SFT & Thinking end-to-end & 434 & 217 & 332 & 132 & 60.8\\
\midrule
RL & Direct answer & 434 & 217 & 311 & 116 & 53.5\\
RL & Thinking first pass & 340 & 135 & 285 & 98 & 45.2\\
RL & Thinking end-to-end & 434 & 217 & 339 & 138 & 63.6\\
\bottomrule
\end{tabular}
\end{table*}

Continuation raises RL robust accuracy to 138/217 (63.6\%) and SFT to 132/217 (60.8\%), restoring 40 and 37 jointly correct pairs. The RL recovery is 18.4 points (bootstrap interval, 13.4--23.5), with all accepted first-pass answers retained.

The RL first pass supplies both answers for 135/217 pairs (62.2\%) and is jointly correct on 98/135 of them (72.6\%). Their product gives the unconditional score exactly: $(135/217)(98/135)=98/217=45.2\%$. For SFT, 133/217 pairs are available and 95/133 are correct, giving 43.8\% overall. Thus conditional correctness and unconditional availability describe different aspects of the same generations.

\subsection{Repetition accompanies failed answer completion}
The failure pattern is already present before clinical RL: SFT fails to supply an accepted first-pass answer in 93/434 cases (21.4\%), compared with 94/434 (21.7\%) after RL. Most failed outputs remain in the thought channel near the cap; Appendix~\ref{app:tokenizer} and Table~\ref{tab:failure_signatures} give the stopping signatures. The similar rates place unfinished reasoning in both adaptation stages.

Repeated selections also occur in the SFT traces: 26/75 extractable failures repeat an explicit commitment. The behavior therefore precedes the RL reward used in this study.

Figure~\ref{fig:budget} shows token lengths and prefix availability. RL availability reaches 301/434 cases and 105/217 pairs at 1,000 tokens, 332/434 cases and 128/217 pairs at 2,000, and 340/434 cases and 135/217 pairs at 3,000. The last 1,000 tokens add eight accepted cases, a 1.8-point increase.

\begin{figure*}[t]
\centering
\begin{minipage}[t]{.48\textwidth}
\centering
\begin{tikzpicture}
\begin{axis}[paperaxis,width=\linewidth,height=5.0cm,xmode=log,xmin=100,xmax=3200,ymin=0,ymax=1,
 xlabel={Reconstructed first-pass tokens},ylabel={Cumulative fraction},ymajorgrids=true,
 xtick={250,500,1000,2000,3000},xticklabels={250,500,1000,2000,3000},
 legend style={at={(.03,.97)},anchor=north west,font=\scriptsize,draw=none}]
\addplot[const plot,no marks,color=ink,line width=1pt] coordinates {(284,0.002304) (286,0.004608) (301,0.006912) (339,0.009217) (349,0.011521) (353,0.013825) (359,0.016129) (360,0.018433) (363,0.020737) (363,0.023041) (369,0.025346) (371,0.027650) (373,0.029954) (389,0.032258) (391,0.034562) (392,0.036866) (393,0.039171) (394,0.041475) (396,0.043779) (396,0.046083) (397,0.048387) (398,0.050691) (399,0.052995) (400,0.055300) (401,0.057604) (402,0.059908) (414,0.062212) (418,0.064516) (422,0.066820) (427,0.069124) (429,0.071429) (432,0.073733) (433,0.076037) (434,0.078341) (434,0.080645) (444,0.082949) (445,0.085253) (447,0.087558) (451,0.089862) (452,0.092166) (453,0.094470) (453,0.096774) (456,0.099078) (458,0.101382) (459,0.103687) (459,0.105991) (461,0.108295) (461,0.110599) (461,0.112903) (462,0.115207) (462,0.117512) (462,0.119816) (463,0.122120) (464,0.124424) (464,0.126728) (464,0.129032) (465,0.131336) (465,0.133641) (466,0.135945) (470,0.138249) (470,0.140553) (471,0.142857) (475,0.145161) (475,0.147465) (475,0.149770) (476,0.152074) (477,0.154378) (477,0.156682) (477,0.158986) (478,0.161290) (478,0.163594) (478,0.165899) (478,0.168203) (479,0.170507) (480,0.172811) (480,0.175115) (481,0.177419) (484,0.179724) (485,0.182028) (485,0.184332) (485,0.186636) (486,0.188940) (487,0.191244) (488,0.193548) (488,0.195853) (488,0.198157) (494,0.200461) (495,0.202765) (496,0.205069) (502,0.207373) (502,0.209677) (503,0.211982) (503,0.214286) (507,0.216590) (509,0.218894) (509,0.221198) (512,0.223502) (513,0.225806) (514,0.228111) (514,0.230415) (516,0.232719) (517,0.235023) (518,0.237327) (518,0.239631) (519,0.241935) (520,0.244240) (521,0.246544) (523,0.248848) (523,0.251152) (524,0.253456) (524,0.255760) (524,0.258065) (524,0.260369) (525,0.262673) (526,0.264977) (526,0.267281) (528,0.269585) (529,0.271889) (529,0.274194) (530,0.276498) (530,0.278802) (532,0.281106) (533,0.283410) (534,0.285714) (536,0.288018) (536,0.290323) (536,0.292627) (536,0.294931) (536,0.297235) (537,0.299539) (537,0.301843) (537,0.304147) (538,0.306452) (539,0.308756) (539,0.311060) (540,0.313364) (543,0.315668) (543,0.317972) (545,0.320276) (545,0.322581) (546,0.324885) (546,0.327189) (546,0.329493) (547,0.331797) (548,0.334101) (549,0.336406) (550,0.338710) (550,0.341014) (551,0.343318) (553,0.345622) (553,0.347926) (554,0.350230) (554,0.352535) (557,0.354839) (557,0.357143) (559,0.359447) (560,0.361751) (560,0.364055) (561,0.366359) (561,0.368664) (561,0.370968) (561,0.373272) (562,0.375576) (562,0.377880) (562,0.380184) (563,0.382488) (564,0.384793) (564,0.387097) (565,0.389401) (567,0.391705) (568,0.394009) (570,0.396313) (570,0.398618) (572,0.400922) (572,0.403226) (572,0.405530) (573,0.407834) (573,0.410138) (575,0.412442) (576,0.414747) (579,0.417051) (579,0.419355) (580,0.421659) (580,0.423963) (580,0.426267) (581,0.428571) (581,0.430876) (583,0.433180) (585,0.435484) (587,0.437788) (588,0.440092) (589,0.442396) (590,0.444700) (591,0.447005) (595,0.449309) (596,0.451613) (597,0.453917) (597,0.456221) (598,0.458525) (599,0.460829) (600,0.463134) (601,0.465438) (601,0.467742) (604,0.470046) (604,0.472350) (604,0.474654) (604,0.476959) (605,0.479263) (606,0.481567) (606,0.483871) (610,0.486175) (612,0.488479) (613,0.490783) (613,0.493088) (616,0.495392) (618,0.497696) (619,0.500000) (619,0.502304) (619,0.504608) (620,0.506912) (622,0.509217) (622,0.511521) (623,0.513825) (623,0.516129) (624,0.518433) (624,0.520737) (627,0.523041) (630,0.525346) (631,0.527650) (632,0.529954) (634,0.532258) (636,0.534562) (638,0.536866) (644,0.539171) (649,0.541475) (649,0.543779) (653,0.546083) (654,0.548387) (656,0.550691) (657,0.552995) (662,0.555300) (662,0.557604) (663,0.559908) (664,0.562212) (665,0.564516) (666,0.566820) (669,0.569124) (670,0.571429) (681,0.573733) (683,0.576037) (684,0.578341) (685,0.580645) (685,0.582949) (688,0.585253) (699,0.587558) (699,0.589862) (701,0.592166) (702,0.594470) (704,0.596774) (710,0.599078) (712,0.601382) (714,0.603687) (715,0.605991) (716,0.608295) (718,0.610599) (718,0.612903) (718,0.615207) (720,0.617512) (721,0.619816) (721,0.622120) (734,0.624424) (735,0.626728) (749,0.629032) (751,0.631336) (755,0.633641) (768,0.635945) (769,0.638249) (774,0.640553) (782,0.642857) (785,0.645161) (807,0.647465) (810,0.649770) (817,0.652074) (819,0.654378) (819,0.656682) (819,0.658986) (822,0.661290) (857,0.663594) (858,0.665899) (889,0.668203) (891,0.670507) (897,0.672811) (901,0.675115) (909,0.677419) (911,0.679724) (913,0.682028) (914,0.684332) (917,0.686636) (924,0.688940) (926,0.691244) (927,0.693548) (929,0.695853) (939,0.698157) (958,0.700461) (977,0.702765) (989,0.705069) (1002,0.707373) (1007,0.709677) (1035,0.711982) (1037,0.714286) (1076,0.716590) (1093,0.718894) (1094,0.721198) (1139,0.723502) (1162,0.725806) (1193,0.728111) (1209,0.730415) (1284,0.732719) (1317,0.735023) (1501,0.737327) (1551,0.739631) (1564,0.741935) (1626,0.744240) (1716,0.746544) (1818,0.748848) (1853,0.751152) (2113,0.753456) (2259,0.755760) (2366,0.758065) (2762,0.760369) (3000,0.762673) (3000,0.764977) (3000,0.767281) (3000,0.769585) (3000,0.771889) (3000,0.774194) (3000,0.776498) (3000,0.778802) (3000,0.781106) (3000,0.783410) (3000,0.785714) (3000,0.788018) (3000,0.790323) (3000,0.792627) (3000,0.794931) (3000,0.797235) (3000,0.799539) (3000,0.801843) (3000,0.804147) (3000,0.806452) (3000,0.808756) (3000,0.811060) (3000,0.813364) (3000,0.815668) (3000,0.817972) (3000,0.820276) (3000,0.822581) (3000,0.824885) (3000,0.827189) (3000,0.829493) (3000,0.831797) (3000,0.834101) (3000,0.836406) (3000,0.838710) (3000,0.841014) (3000,0.843318) (3000,0.845622) (3000,0.847926) (3000,0.850230) (3000,0.852535) (3000,0.854839) (3000,0.857143) (3000,0.859447) (3000,0.861751) (3000,0.864055) (3000,0.866359) (3000,0.868664) (3000,0.870968) (3000,0.873272) (3000,0.875576) (3000,0.877880) (3000,0.880184) (3000,0.882488) (3000,0.884793) (3000,0.887097) (3000,0.889401) (3000,0.891705) (3000,0.894009) (3000,0.896313) (3000,0.898618) (3000,0.900922) (3000,0.903226) (3000,0.905530) (3000,0.907834) (3000,0.910138) (3000,0.912442) (3000,0.914747) (3000,0.917051) (3000,0.919355) (3000,0.921659) (3000,0.923963) (3000,0.926267) (3000,0.928571) (3000,0.930876) (3000,0.933180) (3000,0.935484) (3000,0.937788) (3000,0.940092) (3000,0.942396) (3000,0.944700) (3000,0.947005) (3000,0.949309) (3000,0.951613) (3000,0.953917) (3000,0.956221) (3000,0.958525) (3000,0.960829) (3000,0.963134) (3000,0.965438) (3000,0.967742) (3000,0.970046) (3000,0.972350) (3000,0.974654) (3000,0.976959) (3000,0.979263) (3000,0.981567) (3000,0.983871) (3000,0.986175) (3000,0.988479) (3000,0.990783) (3000,0.993088) (3000,0.995392) (3000,0.997696) (3000,1.000000)};
\addplot[const plot,no marks,color=evidence,line width=1pt] coordinates {(307,0.002304) (314,0.004608) (332,0.006912) (334,0.009217) (349,0.011521) (355,0.013825) (360,0.016129) (363,0.018433) (367,0.020737) (369,0.023041) (372,0.025346) (374,0.027650) (375,0.029954) (381,0.032258) (386,0.034562) (386,0.036866) (387,0.039171) (387,0.041475) (392,0.043779) (396,0.046083) (396,0.048387) (396,0.050691) (397,0.052995) (401,0.055300) (403,0.057604) (404,0.059908) (404,0.062212) (408,0.064516) (410,0.066820) (413,0.069124) (415,0.071429) (418,0.073733) (419,0.076037) (422,0.078341) (426,0.080645) (426,0.082949) (429,0.085253) (431,0.087558) (432,0.089862) (432,0.092166) (432,0.094470) (435,0.096774) (436,0.099078) (437,0.101382) (439,0.103687) (440,0.105991) (444,0.108295) (444,0.110599) (444,0.112903) (455,0.115207) (456,0.117512) (457,0.119816) (458,0.122120) (459,0.124424) (459,0.126728) (459,0.129032) (461,0.131336) (462,0.133641) (462,0.135945) (463,0.138249) (464,0.140553) (465,0.142857) (465,0.145161) (468,0.147465) (471,0.149770) (471,0.152074) (472,0.154378) (473,0.156682) (473,0.158986) (473,0.161290) (473,0.163594) (475,0.165899) (475,0.168203) (475,0.170507) (475,0.172811) (475,0.175115) (477,0.177419) (478,0.179724) (479,0.182028) (480,0.184332) (481,0.186636) (482,0.188940) (482,0.191244) (486,0.193548) (486,0.195853) (487,0.198157) (488,0.200461) (492,0.202765) (492,0.205069) (493,0.207373) (493,0.209677) (495,0.211982) (495,0.214286) (496,0.216590) (497,0.218894) (501,0.221198) (506,0.223502) (508,0.225806) (511,0.228111) (512,0.230415) (513,0.232719) (515,0.235023) (515,0.237327) (516,0.239631) (516,0.241935) (519,0.244240) (519,0.246544) (520,0.248848) (520,0.251152) (523,0.253456) (523,0.255760) (523,0.258065) (524,0.260369) (526,0.262673) (527,0.264977) (528,0.267281) (529,0.269585) (529,0.271889) (530,0.274194) (530,0.276498) (530,0.278802) (530,0.281106) (530,0.283410) (534,0.285714) (534,0.288018) (535,0.290323) (537,0.292627) (538,0.294931) (540,0.297235) (541,0.299539) (541,0.301843) (541,0.304147) (542,0.306452) (544,0.308756) (544,0.311060) (544,0.313364) (544,0.315668) (545,0.317972) (545,0.320276) (547,0.322581) (547,0.324885) (547,0.327189) (548,0.329493) (548,0.331797) (549,0.334101) (549,0.336406) (550,0.338710) (551,0.341014) (551,0.343318) (556,0.345622) (556,0.347926) (556,0.350230) (556,0.352535) (556,0.354839) (558,0.357143) (558,0.359447) (559,0.361751) (563,0.364055) (563,0.366359) (564,0.368664) (564,0.370968) (564,0.373272) (565,0.375576) (565,0.377880) (565,0.380184) (567,0.382488) (567,0.384793) (568,0.387097) (569,0.389401) (569,0.391705) (570,0.394009) (571,0.396313) (573,0.398618) (573,0.400922) (575,0.403226) (575,0.405530) (575,0.407834) (576,0.410138) (577,0.412442) (577,0.414747) (577,0.417051) (578,0.419355) (578,0.421659) (579,0.423963) (579,0.426267) (580,0.428571) (582,0.430876) (583,0.433180) (583,0.435484) (585,0.437788) (585,0.440092) (585,0.442396) (586,0.444700) (587,0.447005) (588,0.449309) (588,0.451613) (590,0.453917) (592,0.456221) (592,0.458525) (595,0.460829) (595,0.463134) (596,0.465438) (596,0.467742) (598,0.470046) (599,0.472350) (602,0.474654) (603,0.476959) (605,0.479263) (612,0.481567) (613,0.483871) (613,0.486175) (613,0.488479) (615,0.490783) (617,0.493088) (620,0.495392) (622,0.497696) (624,0.500000) (625,0.502304) (625,0.504608) (625,0.506912) (625,0.509217) (626,0.511521) (627,0.513825) (627,0.516129) (629,0.518433) (633,0.520737) (633,0.523041) (633,0.525346) (636,0.527650) (638,0.529954) (638,0.532258) (639,0.534562) (641,0.536866) (642,0.539171) (642,0.541475) (648,0.543779) (653,0.546083) (656,0.548387) (658,0.550691) (659,0.552995) (659,0.555300) (663,0.557604) (664,0.559908) (667,0.562212) (671,0.564516) (679,0.566820) (683,0.569124) (683,0.571429) (684,0.573733) (686,0.576037) (690,0.578341) (692,0.580645) (695,0.582949) (699,0.585253) (700,0.587558) (700,0.589862) (703,0.592166) (703,0.594470) (705,0.596774) (707,0.599078) (712,0.601382) (717,0.603687) (719,0.605991) (730,0.608295) (744,0.610599) (762,0.612903) (763,0.615207) (765,0.617512) (765,0.619816) (771,0.622120) (772,0.624424) (773,0.626728) (773,0.629032) (774,0.631336) (775,0.633641) (790,0.635945) (791,0.638249) (791,0.640553) (791,0.642857) (802,0.645161) (804,0.647465) (805,0.649770) (806,0.652074) (855,0.654378) (855,0.656682) (863,0.658986) (863,0.661290) (865,0.663594) (875,0.665899) (879,0.668203) (894,0.670507) (906,0.672811) (928,0.675115) (932,0.677419) (939,0.679724) (962,0.682028) (963,0.684332) (972,0.686636) (974,0.688940) (977,0.691244) (1005,0.693548) (1021,0.695853) (1033,0.698157) (1068,0.700461) (1099,0.702765) (1105,0.705069) (1119,0.707373) (1123,0.709677) (1138,0.711982) (1257,0.714286) (1259,0.716590) (1362,0.718894) (1468,0.721198) (1516,0.723502) (1570,0.725806) (1923,0.728111) (2152,0.730415) (2166,0.732719) (2593,0.735023) (2647,0.737327) (3000,0.739631) (3000,0.741935) (3000,0.744240) (3000,0.746544) (3000,0.748848) (3000,0.751152) (3000,0.753456) (3000,0.755760) (3000,0.758065) (3000,0.760369) (3000,0.762673) (3000,0.764977) (3000,0.767281) (3000,0.769585) (3000,0.771889) (3000,0.774194) (3000,0.776498) (3000,0.778802) (3000,0.781106) (3000,0.783410) (3000,0.785714) (3000,0.788018) (3000,0.790323) (3000,0.792627) (3000,0.794931) (3000,0.797235) (3000,0.799539) (3000,0.801843) (3000,0.804147) (3000,0.806452) (3000,0.808756) (3000,0.811060) (3000,0.813364) (3000,0.815668) (3000,0.817972) (3000,0.820276) (3000,0.822581) (3000,0.824885) (3000,0.827189) (3000,0.829493) (3000,0.831797) (3000,0.834101) (3000,0.836406) (3000,0.838710) (3000,0.841014) (3000,0.843318) (3000,0.845622) (3000,0.847926) (3000,0.850230) (3000,0.852535) (3000,0.854839) (3000,0.857143) (3000,0.859447) (3000,0.861751) (3000,0.864055) (3000,0.866359) (3000,0.868664) (3000,0.870968) (3000,0.873272) (3000,0.875576) (3000,0.877880) (3000,0.880184) (3000,0.882488) (3000,0.884793) (3000,0.887097) (3000,0.889401) (3000,0.891705) (3000,0.894009) (3000,0.896313) (3000,0.898618) (3000,0.900922) (3000,0.903226) (3000,0.905530) (3000,0.907834) (3000,0.910138) (3000,0.912442) (3000,0.914747) (3000,0.917051) (3000,0.919355) (3000,0.921659) (3000,0.923963) (3000,0.926267) (3000,0.928571) (3000,0.930876) (3000,0.933180) (3000,0.935484) (3000,0.937788) (3000,0.940092) (3000,0.942396) (3000,0.944700) (3000,0.947005) (3000,0.949309) (3000,0.951613) (3000,0.953917) (3000,0.956221) (3000,0.958525) (3000,0.960829) (3000,0.963134) (3000,0.965438) (3000,0.967742) (3000,0.970046) (3000,0.972350) (3000,0.974654) (3000,0.976959) (3000,0.979263) (3000,0.981567) (3000,0.983871) (3000,0.986175) (3000,0.988479) (3000,0.990783) (3000,0.993088) (3000,0.995392) (3001,0.997696) (3001,1.000000)};
\legend{SFT,RL}
\end{axis}
\end{tikzpicture}
\end{minipage}\hfill
\begin{minipage}[t]{.48\textwidth}
\centering
\begin{tikzpicture}
\begin{axis}[paperaxis,width=\linewidth,height=5.0cm,xmin=250,xmax=3000,ymin=0,ymax=100,
 xlabel={Saved-prefix token limit},ylabel={Available (\%)},ymajorgrids=true,
 xtick={500,1000,2000,3000},legend style={at={(.97,.03)},anchor=south east,font=\scriptsize,draw=none}]
\addplot[color=ink,mark=*,mark size=1pt] coordinates {(250,0.00000) (500,21.19816) (750,63.13364) (1000,71.19816) (1500,75.80645) (2000,78.11060) (2500,78.11060) (2750,78.11060) (3000,78.57143)};
\addplot[color=evidence,mark=*,mark size=1pt] coordinates {(250,0.00000) (500,22.35023) (750,61.52074) (1000,69.35484) (1500,73.96313) (2000,76.49770) (2500,78.11060) (2750,78.34101) (3000,78.34101)};
\addplot[color=ink,dashed,mark=none] coordinates {(250,0.00000) (500,6.45161) (750,42.85714) (1000,50.23041) (1500,56.68203) (2000,60.82949) (2500,60.82949) (2750,60.82949) (3000,61.29032)};
\addplot[color=evidence,dashed,mark=none] coordinates {(250,0.00000) (500,6.45161) (750,39.63134) (1000,48.38710) (1500,55.76037) (2000,58.98618) (2500,61.75115) (2750,62.21198) (3000,62.21198)};
\legend{SFT cases,RL cases,SFT pairs,RL pairs}
\end{axis}
\end{tikzpicture}
\end{minipage}
\caption{Length and availability in fixed first-pass trajectories. Left: empirical token-length distributions. Right: parser availability at saved prefixes, with denominators of 434 cases and 217 pairs. RL case availability rises from 332 to 340 between 2,000 and 3,000 tokens.}
\label{fig:budget}
\end{figure*}

Conditional yields decline within the principal run (Table~\ref{tab:interval_yields}, Appendix~\ref{app:secondary}): 31/133 previously unavailable cases gain an answer between 1,000 and 2,000 tokens (23.3\%), followed by 8/102 between 2,000 and 3,000 (7.8\%). Excluding the one output already stopped below 2,000 gives an active-trace yield of 8/101 (7.9\%).

A longer saved configuration also has limited late answer yield: 17/204 unanswered traces extending beyond 3,000 tokens supply an accepted answer by 4,096. Its configuration and within-run yields are reported in Appendix~\ref{app:secondary}.

The extracted RL diagnosis recurs at least twice in 71/75 final thought windows (94.7\%; median 10 mentions, IQR 7--13.5). Yet the accepted-answer control also exceeds this threshold in 338/340 cases (99.4\%; median 5, IQR 4--7). Using the continuation diagnosis in all failures gives 89/94 (94.7\%; median 9.5, IQR 7--12). This threshold captures routine diagnostic discussion as well as unfinished reasoning (Table~\ref{tab:recurrence_control}).

\begin{table}[t]
\centering\scriptsize
\caption{RL recurrence in the final 500 thought tokens. Mentions and assertions count traces with at least two matches. Assertions use frozen selection rules. The extracted-commitment group contains the 75 eligible failures; the continuation-target group retains all 94.}
\label{tab:recurrence_control}
\begin{tabularx}{\linewidth}{@{}Xrrr@{}}
\toprule
Group and diagnosis target & $n$ & Mentions & Assertions\\
\midrule
Available; accepted diagnosis & 340 & 338 & 15\\
Failed; extracted commitment & 75 & 71 & 35\\
Failed; continuation diagnosis & 94 & 89 & 33\\
\bottomrule
\end{tabularx}
\end{table}

Explicit answer selection gives a sharper contrast: repeated assertions occur in 35/75 extractable failures (46.7\%), versus 15/340 available first passes (4.4\%). With the continuation diagnosis as target, the rate is 33/94 failures (35.1\%). Four extracted diagnoses occur before the final window and have zero mentions there; none has exactly one. The extracted-target analysis abstains on the other 19 failures.

Whole-trace recurrence is also higher in the rescue group: median $r_4$ is 0.491--0.503 for rescued RL traces and 0.047--0.055 for unrescued traces (Table~\ref{tab:repetition_stratified}). This statistic counts duplicate word windows across the full thought, whereas the exact-cycle detector requires adjacent identical token blocks. Figure~\ref{fig:rehearsal} (Appendix~\ref{app:traces}) illustrates explicit reassertion in one unfinished trace.

\subsection{Reading commitments recovers paired correctness}
For the RL fallback, 67 matching diagnoses inherit continuation verdicts and three use unambiguous archived verdicts for the same case. Of the five newly adjudicated diagnoses, two are accepted. Across all 75 candidates, 41 are accepted and 34 rejected.

The extraction-only fallback supplies candidates for 415/434 cases and both members of 199/217 pairs. It yields 129/217 jointly correct pairs (59.4\%). Against direct answering, there are 27 gains and 14 losses: $+6.0$ points (paired-bootstrap 95\% interval, $+0.5$ to $+11.5$; exact McNemar $p=0.0596$). The comparison remains borderline under the exact test on 41 discordant pairs.

The fallback restores 30 of the 40 pairs recovered by continuation. In 29, every rescued member has the same extracted and continued diagnosis; their matching verdicts are reused. The fallback also restores one pair that continuation misses. Figure~\ref{fig:commitment} shows how continuation can preserve a diagnosis, revise it, or lose a scored qualifier.

\begin{figure*}[t]
\centering
\begin{minipage}[t]{.32\textwidth}
\tracepanel{evidence}{Already asserted\hfill CASE\_003}{%
First pass: \textit{``Let's stick with vasa previa''}.\par\smallskip
Continuation: \textbf{vasa previa}.\par\smallskip
The existing verdict accepts this diagnosis. A matching explicit assertion is available before continuation.}
\end{minipage}\hfill
\begin{minipage}[t]{.32\textwidth}
\tracepanel{ink}{Changed diagnosis\hfill CASE\_415}{%
First pass: \textit{``Let's go with Juvenile Idiopathic Arthritis''}.\par\smallskip
Continuation: \textbf{osteosarcoma}.\par\smallskip
The archived verdict rejects the first diagnosis and accepts the continuation diagnosis.}
\end{minipage}\hfill
\begin{minipage}[t]{.32\textwidth}
\tracepanel{rehearsal}{Lost specificity\hfill CASE\_159}{%
First pass: \textit{``the most likely diagnosis is postpartum hemorrhage due to coagulopathy''}.\par\smallskip
Continuation: \textbf{postpartum hemorrhage}.\par\smallskip
The specific assertion has an accepted archived verdict; the broader continuation prediction is rejected.}
\end{minipage}
\caption{Diagnosis preservation, revision, and qualifier loss in saved first passes and continuations. Quoted assertions are exact first-pass spans; continuation diagnoses are parsed predictions. Verdicts come from the same-case archived labels.}
\label{fig:commitment}
\end{figure*}

The examples link readout to clinical specificity. CASE\_003 preserves vasa previa, CASE\_415 changes the diagnosis from juvenile idiopathic arthritis to osteosarcoma, and CASE\_159 drops the coagulopathy qualifier from postpartum hemorrhage. The archived verdicts respectively record preservation, improvement, and loss of an accepted diagnosis.

\subsection{Frozen-extractor transfer to SFT}
For SFT, 58 extracted diagnoses match the continuation, one has an unambiguous archived verdict, and one exactly matches the reference. Of 15 newly adjudicated candidates, five are accepted. Across the 75 extracted diagnoses, 42 are accepted and 33 rejected. The fallback supplies 416/434 cases and both members of 199/217 pairs, yielding 123/217 jointly correct pairs (56.7\%). It restores 28 of the 37 continuation-recovered pairs.

Table~\ref{tab:fallback_transfer} reports the completed evaluation. SFT has 26 gains and 16 losses against direct answering: $+4.6$ points (paired-bootstrap interval, $-1.4$ to $+10.6$; McNemar $p=0.1641$). The frozen rules transfer extraction and within-trace recovery to SFT, while its comparison with direct answering remains inconclusive.

\begin{table*}[t]
\centering\small
\caption{Completed fallback evaluation after adjudicating all extracted diagnoses. Every score uses 217 pairs. Differences and paired-bootstrap 95\% intervals are in percentage points, relative to the corresponding direct-answer condition. $p$ is the exact McNemar result.}
\label{tab:fallback_transfer}
\setlength{\tabcolsep}{5pt}
\begin{tabular}{@{}lrrrrrr@{}}
\toprule
Checkpoint & Correct pairs / 217 & Robust (\%) & Difference & 95\% interval & Gains / losses & $p$\\
\midrule
RL & 129 & 59.4 & $+6.0$ & $[+0.5,+11.5]$ & 27 / 14 & 0.0596\\
SFT & 123 & 56.7 & $+4.6$ & $[-1.4,+10.6]$ & 26 / 16 & 0.1641\\
\bottomrule
\end{tabular}
\end{table*}

\subsection{A second commitment supports an earlier readout}
The prefix-only trigger fires in 91 RL cases, including 59 original first-pass failures, and in 72 SFT cases, including 49 failures (Table~\ref{tab:early_readout}). The selected diagnosis matches the saved final prediction in 84/91 RL triggers and 67/72 SFT triggers. Median trigger positions are 1,266 and 1,381 reconstructed tokens, respectively.

Retaining these prefixes omits 24.1\% of RL and 18.5\% of SFT first-pass text across all cases. The resulting readout reaches 122/217 jointly correct RL pairs (56.2\%) and 114/217 SFT pairs (52.5\%), compared with 59.4\% and 56.7\% when reading the last eligible commitment at the observed end. This quantifies an earlier-readout tradeoff: repeated selection can expose a usable diagnosis sooner, while some later diagnostic changes improve paired correctness.

\begin{table*}[t]
\centering\small
\caption{Offline second-commitment readout on all 434 cases per checkpoint. Agreement compares triggered diagnoses with saved final predictions. Omitted text is measured over reconstructed first-pass tokens; scores use 217 pairs and completed diagnostic labels.}
\label{tab:early_readout}
\begin{tabular}{@{}lrrrrr@{}}
\toprule
Checkpoint & Triggered cases & Diagnosis matches & Omitted text (\%) & Correct pairs / 217 & Robust (\%)\\
\midrule
RL & 91 & 84 / 91 & 24.1 & 122 & 56.2\\
SFT & 72 & 67 / 72 & 18.5 & 114 & 52.5\\
\bottomrule
\end{tabular}
\end{table*}

\section{Discussion}
The main result concerns diagnostic content: in most extractable failed first passes, continuation reproduces the last explicitly asserted diagnosis. This explains why a missing structured answer can coexist with a recoverable clinical conclusion. Under the saved labels, reading those assertions alone restores three quarters of the RL pairs recovered by continuation.

Continuation also changes some answers. The trace examples show both a beneficial revision and a loss of diagnostic specificity. Clinical answer recovery therefore involves preserving the selected diagnosis and its consequential qualifiers, alongside satisfying the output schema. This connects the clinical evaluation problem to intermediate-answer analysis and overthinking \cite{measuringfaithfulness,overthinking}, as well as inference control and structured generation \cite{s1,tokenbudget,budgetthinker,testcompute,picard,guided}.

The near-identical SFT and RL failure rates show that unfinished answers are already present at the supervised stage. The recurrence controls clarify what is informative in a reasoning trace. Repeated mentions of a candidate diagnosis are routine among accepted outputs; repeated explicit selections are more concentrated in the failed group. Together with diminishing late answer yield, this suggests a practical direction for inference design: monitor repeated commitment statements and test whether earlier answer emission preserves diagnostic specificity.

The offline trigger makes this direction concrete: a second matching assertion exposes an earlier readout, with a measurable text--accuracy tradeoff. A live test can now compare direct extraction with forced answer continuation at the same trigger.

For evaluation, the useful decomposition has three parts: the diagnosis stated in thought, the answer accepted by the interface, and correctness on both clinical presentations. Reporting each part makes it possible to locate where a continuation changes an outcome. The fallback's borderline RL comparison and inconclusive SFT comparison with direct answering are secondary to this within-trace diagnosis-preservation finding.

\section{Limitations}
This study covers one Gemma 4 31B lineage, one RL run, and one benchmark. SFT and RL share cases and generation priors; extraction rules were developed on the RL failures before being frozen for SFT. Direct-answer and thinking conditions differ in budget and control, and the saved records lack runtime finish reasons. Prefix yields and trigger reductions describe the observed text; they do not measure runtime savings or the answers a fresh continuation would produce. Recurrence comparisons also vary in trace length, target selection, and extraction eligibility.

Correctness scores reuse archived verdicts. Additional extracted diagnoses are scored with the original evaluator, and residual disagreements across scoring snapshots affect small between-checkpoint differences (Appendix~\ref{app:provenance}). Bootstrap intervals summarize variation over benchmark pairs. The clinician sample contains 35 selected cases, exposes automatic verdicts, and has low agreement on several axes.

Training-data overlap with MamaBench has not been exhaustively checked. Historical SFT template and loss-masking details remain unreconciled, and the RL run uses a frozen SFT rollout policy with clipped importance correction (Appendix~\ref{app:training}). These factors define the scope for checkpoint reproduction and broader clinical validation.

\section{Conclusion}
Continuation often returns a diagnosis already stated in unfinished clinical reasoning: 67/75 extractable RL failures and 58/75 SFT failures preserve that diagnosis in MamaReason. The key finding is agreement in diagnostic content before and after continuation, measured independently of correctness labels. Reading the earlier commitments also recovers a substantial share of paired correctness under the archived verdicts.

Clinical reasoning evaluations should report diagnostic content, answer availability, and paired correctness together. Logging native stopping information and preserving first-pass text would support this decomposition. The next experiments are matched-budget answer-control comparisons, blinded clinical review of diagnosis and qualifier changes, and live tests of the second-commitment trigger.

\clearpage
\appendix
\section{Training and Reward Specification}
\label{app:training}
\subsection{Data preparation}
The SFT source contains 28,082 question--answer records. Lowercasing and trimming questions before exact deduplication leaves 22,798 unique questions. Sampling targets 12,000 questions with seed 42, stratified by heuristic language and topic cues. Language cues include Nigerian Pidgin markers and selected orthographic features; topics include antenatal care, preeclampsia, infection prevention, and access to care.

A teacher generates stepwise reasoning and a concise answer for each sampled question in the Nigerian maternal and child-health context, using temperature 0.3 and a 2,048-token cap. The source answer is withheld during generation. The saved output contains 11,591 conversations. Conversion separates thought and visible-answer channels, and structural cleaning removes off-topic records and missing, empty, or short thought. Removing 936 flagged identifiers leaves 10,655 conversations, split 95:5 with a fixed seed into 10,122 training and 533 validation examples. Filtering checks structure and length; semantic answer verification is outside this pipeline.

Checks screen for short or incomplete records, exact duplicate prefixes, and possible diagnosis leakage. Condition cleaning and metadata re-extraction yield 1,627 vignettes across 111 normalized conditions. The RL run uses a 1,000-vignette subset sampled with seed 42 by proportional urgency-by-setting allocation and a largest-remainder rule: 571 emergent, 397 urgent, and 32 routine cases. It retains all 111 conditions and is used without a validation split. Table~\ref{tab:data} summarizes the retained preparation stages.

\begin{table}[t]
\centering\small
\caption{Data construction. The final 1,000-case subset is the dataset used for RL; the larger pool is not the training size.}
\label{tab:data}
\begin{tabularx}{\linewidth}{@{}Xr@{}}
\toprule
Stage & Records\\
\midrule
Supervised source QA & 28,082\\
Unique normalized questions & 22,798\\
Sampling target & 12,000\\
Saved reasoning demonstrations & 11,591\\
Cleaned demonstrations & 10,655\\
SFT train / validation & 10,122 / 533\\
\midrule
Re-extracted vignette pool & 1,627\\
RL subset (111 conditions) & 1,000\\
Emergent / urgent / routine & 571 / 397 / 32\\
\bottomrule
\end{tabularx}
\end{table}

\subsection{Optimization and reward}

We initialize RL from the SFT checkpoint. Both adaptation stages use four-bit NF4 quantization with double quantization and BF16 computation, with LoRA adapters on non-vision attention and feed-forward projections \cite{qlora,lora}. The saved SFT configuration specifies rank 64, scaling 64, dropout 0.05, two epochs, a 1,024-token sequence cap, and learning rate $5\times10^{-5}$ with cosine decay. RL uses rank 32, scaling 32, the same dropout, one epoch, and learning rate $10^{-5}$. Prompts request clinical reasoning, differential diagnoses, investigations, management, and explicit triage and diagnosis fields.

For each prompt, the RL implementation samples $G=4$ completions at temperature 0.80 and top-$p=0.95$, with prompt and completion caps of 2,048 tokens each. PPO clipping is 0.20 \cite{ppo} and the KL coefficient against the SFT reference is 0.001. The implementation uses TRL's fixed-denominator \texttt{dr\_grpo} reduction with default group-standardized advantages \cite{trl}; the reduction removes sequence-length normalization while retaining group-standardized advantages \cite{drgrpo}. Colocated rollouts use a frozen SFT behavior policy, while the adapter policy is updated, with importance-sampling correction capped at 2.0. The behavior policy remains the SFT policy during these adapter updates.

Let $D$, $F$, $U$, and $M$ denote diagnosis, critical-finding coverage, urgency, and management scores. The four-component implementation normalizes their relative weights to obtain
\begin{equation}
R=\frac{1}{3}D+\frac{5}{18}F+\frac{1}{6}U+\frac{2}{9}M.
\label{eq:reward}
\end{equation}
Diagnosis scoring canonicalizes aliases and applies substring or token-set matching. A primary-reference match or Jaccard overlap of at least 0.50 receives 1; overlap from 0.33 to below 0.50 receives 0.70; a reference-differential match receives 0.40. Other predictions receive zero. This lexical training score is distinct from the stricter diagnostic evaluator used for MamaBench.

The finding score measures recall of reference findings and critical concepts. A target matches when any punctuation-stripped word longer than three characters occurs in the raw response. With red flags present, recall of at least 0.75 receives 1, recall from 0.40 to below 0.75 receives 0.40, and lower recall receives $-1$. Without red flags, the score equals recall; an empty target set receives 1. Urgency scoring assigns 1 for a correct routine/urgent/emergent category, 0.30 for one-level overtriage, and $-1$ for emergent-to-routine undertriage. Other mismatches or unknown categories receive zero.

Management combines action recall $a$, investigation recall $t$, and setting feasibility $s$:
\begin{equation}
M=\operatorname{clip}_{[0,1]}\bigl(0.45\min(1,2a)+0.35t+0.20s\bigr).
\end{equation}
Action and test coverage use lexical matching. Feasibility uses referral and stabilization markers in primary or rural settings and admission, investigation, monitoring, or intravenous-treatment markers elsewhere. These rules measure lexical coverage of selected clinical features.

Diagnosis and urgency are preferentially extracted from the visible answer, with thought-channel fallback. Finding and management rules inspect raw response text, including inline thought when present. There is no explicit repetition or token-length penalty. Repeating already matched content can therefore leave the clinical score unchanged, although token probabilities, KL regularization, truncation, and rollout correction still affect learning. It makes answer closure a useful target for a future reward ablation.

\section{Tokenizer and Readout Reproducibility}
\label{app:tokenizer}
\begin{table}[t]
\centering\small
\caption{First-pass failure signatures. Cap-consistent means within one reconstructed token of 3,000. Shorter exact-cycle failures form a separate group.}
\label{tab:failure_signatures}
\begin{tabularx}{\linewidth}{@{}Xrr@{}}
\toprule
Failed-output pattern & RL & SFT\\
\midrule
All first-pass failures & 94 & 93\\
JSON parse failures & 93 & 91\\
Blank/template predictions & 1 & 2\\
Exactly 3,000 reconstructed tokens & 90 & 90\\
Within one token of cap & 92 & 90\\
Below cap with exact suffix cycle & 2 & 3\\
Closed thought channel & 0 & 0\\
Explicit end-of-turn marker & 0 & 0\\
\bottomrule
\end{tabularx}
\end{table}
Reconstruction uses a public base tokenizer with special tokens retained and none automatically added. Exact text round trips are checked on all 868 principal bounded-thinking first passes.

The cap-consistent group includes 90 RL failures of exactly 3,000 reconstructed tokens and two of 3,001 tokens. Re-encoding can differ at token boundaries. The two shorter RL failures are CASE\_090 (2,152 tokens) and CASE\_194 (1,923 tokens); both have an exact trailing cycle. The SFT counterparts are CASE\_102 (2,366 tokens), CASE\_189 (1,626 tokens), and CASE\_238 (1,501 tokens). In each, a repeated 8--32-token block occurs three consecutive times at the suffix. These suffixes match the detector's pattern definition.

In the historical direct-thinking run, 178/193 failures reconstruct to exactly 4,096 tokens, three to 4,095, and three to 4,097. Eight others reconstruct below 4,095 (799, 1,316, 1,426, 1,614, 2,319, 2,427, 3,144, and 4,000), and one to 4,101. All lack a thought-channel closure. The interval analysis uses these saved lengths and the parser outcomes.

The recurrence control excludes the visible answer from each accepted first pass. Of its 340 outputs, 318 have an observed thought-channel closure; 316/318 of these still satisfy the two-mention threshold. The threshold is saturated in the closed-channel subset as well. All 340 controls have nonempty thought text. The continuation-target comparison retains every failed output and applies the same word-boundary phrase matching without using its correctness verdict.

Token-offset truncation retains the source text up to the last retained token's character offset. All prefix availability counts rerun the same JSON parser and placeholder gate. For unclosed thought, the same JSON recovery procedure is applied to the remaining text.

\subsection{Frozen commitment-extraction rules}
\label{app:extractor}
The following rules specify the extractor used for both checkpoints. Matching is case-insensitive and follows textual order.
\begin{enumerate}
\item Recognize an assertion cue from one of these families:
\begin{itemize}
\item ``let's'' (straight or curly apostrophe), followed by ``go with'', ``choose'', ``answer'', ``use'', ``stick with'', ``settle on'', or ``finalize'';
\item ``I'll'' or ``I will'', followed by ``go with'', ``answer'', ``choose'', ``use'', ``state'', or ``pick'';
\item optionally ``the'', then ``final'', ``most likely'', ``working'', ``primary'', or ``definitive'', followed by ``diagnosis'' and ``is'', a colon, or an equals sign;
\item ``my'' or ``the'', optionally ``final'', then ``answer'' or ``prediction'' and ``is'', a colon, or an equals sign;
\item a quoted JSON prediction key, colon, and opening double quote.
\end{itemize}
\item Capture 2--160 characters after the cue, ending before a newline, period, exclamation mark, question mark, brace, or double quote. Trim whitespace, colons, asterisks, and backticks from the ends, remove an initial ``simply'', and cut at a word-boundary occurrence of ``because'', ``given that'', ``since'', or ``for the final answer''.
\item Exclude empty candidates, candidates containing an opening angle bracket or ``condition name'', and candidates beginning with ``that'', ``this'', ``it'', ``still'', ``a more'', ``the same'', ``the best'', ``the most'', ``as above'', ``as the'', ``simply that'', or ``simply this''. Also exclude a match ending at the end of saved text or immediately before a question mark.
\item Select the last remaining candidate by text position. With none, abstain. JSON and prose cues share this ordering.
\end{enumerate}
For scoring and string comparison, lowercase, replace each run outside a--z and 0--9 with a space, and trim the result. \subsection{Additional diagnostic adjudication}
\label{app:adjudication}
The earlier analysis left five RL and 15 SFT last-commitment candidates unscored. These 20 case--diagnosis pairs were subsequently evaluated using the original diagnostic prompt and model, temperature 0, and a maximum of 1,024 response tokens. The evaluator receives the reference diagnosis and candidate prediction, without the reasoning trace. Table~\ref{tab:added_adjudications} lists the candidates and references. All 20 requests returned valid verdicts: two RL candidates and five SFT candidates were accepted. The original benchmark predictions and verdicts remain unchanged.

For archival comparison, the pre-adjudication lower--upper ranges were 129--131 RL pairs (59.4--60.4\%) and 119--129 SFT pairs (54.8--59.4\%). The completed results are 129 and 123 pairs. The former conservative comparisons were $+6.0$ points for RL ($p=0.0596$) and $+2.8$ for SFT ($p=0.4514$); current SFT results are $+4.6$ points ($p=0.1641$). The historical ranges reflected missing candidate verdicts, rather than sampling uncertainty.

The early-trigger readout introduced three further diagnosis strings without an unambiguous saved verdict. All three were scored by the same procedure and accepted. The early-readout scores therefore contain no unresolved diagnostic labels. An event-based implementation was checked against every-token prefix replay on eight representative triggered and untriggered cases across both checkpoints, with matching trigger positions and diagnoses.

\begin{table*}[t]
\centering\scriptsize
\caption{Added diagnostic evaluations. L denotes a last-commitment candidate; E denotes an early-trigger candidate. Verdicts use the same automatic diagnostic criterion as the main evaluation.}
\label{tab:added_adjudications}
\begin{tabularx}{\textwidth}{@{}llcXXc@{}}
\toprule
Checkpoint & Case & Readout & Candidate diagnosis & Reference diagnosis & Correct\\
\midrule
RL & CASE\_006 & L & AFLP & Gall stone & No\\
RL & CASE\_123 & L & varicella & Chicken pox & Yes\\
RL & CASE\_258 & L & GERD & Dyspepsia in pregnancy & No\\
RL & CASE\_342 & L & genital tract trauma causing postpartum hemorrhage & Genital Tract Trauma & Yes\\
RL & CASE\_382 & L & SMA Type III & Muscular Dystrophy & No\\
SFT & CASE\_054 & L & endometrial cancer or endometrial hyperplasia & Endometrial polyps & No\\
SFT & CASE\_120 & L & Neurofibromatosis Type 2 (NF2) & Neurofibromatosis 2 & Yes\\
SFT & CASE\_123 & L & varicella & Chicken pox & Yes\\
SFT & CASE\_134 & L & GBS & Guillain Barre syndrome & Yes\\
SFT & CASE\_135 & L & Global Developmental Delay as it's a widely recognized term in pediatric developmental screening & Mental retardation & Yes\\
SFT & CASE\_169 & L & indeed related to indoor air pollution & Laryngitis & No\\
SFT & CASE\_194 & L & often the one that explains all the symptoms & Germ cell tumor in the pineal region & No\\
SFT & CASE\_248 & L & physiological urinary frequency & Normal pregnancy & No\\
SFT & CASE\_258 & L & preeclampsia & Dyspepsia in pregnancy & No\\
SFT & CASE\_267 & L & Bacterial meningitis (given the constellation of symptoms in an infant) & Febrile seizures & No\\
SFT & CASE\_295 & L & the cause of the bleeding & PPH from Uterine atony & No\\
SFT & CASE\_325 & L & an innocent murmur, specifically Still's murmur & Innocent murmur & Yes\\
SFT & CASE\_352 & L & a benign ovarian cyst & Benign Ovarian Tumour (Serous Cystadenoma) & No\\
SFT & CASE\_382 & L & SMA Type III & Muscular Dystrophy & No\\
SFT & CASE\_421 & L & something like & Cervical Polyp & No\\
RL & CASE\_303 & E & a normal intrauterine pregnancy & Normal pregnancy & Yes\\
RL & CASE\_311 & E & an innocent murmur & Physiologic murmur & Yes\\
SFT & CASE\_365 & E & Hirschsprung disease as it's a classic board-exam presentation for this scenario & Hirschsprung's disease & Yes\\
\bottomrule
\end{tabularx}
\end{table*}

\section{Completed Clinician Review}
\subsection{Review procedure}
Two clinicians complete the same eight-axis workbook for 35 selected RL cases from 25 pairs, yielding 70 case reviews. Axes cover agreement with the automatic verdict, overthinking, underthinking, missed critical information, misweighted findings, hallucinated findings, thought--answer consistency, and five-point clinical acceptability. The automatic verdict remains visible in the workbook.

We join reviews by case identifier and retain each clinician's ratings. Binary axes use raw agreement and Cohen's $\kappa$ \cite{cohen}; acceptability uses rating distributions, exact agreement, and linearly weighted $\kappa$ \cite{weightedkappa}. The following tables describe the selected sample.

\subsection{Review sample and agreement}
\label{app:doctor}
Both completed workbooks contain all eight ratings for the same 35 case identifiers. Predictions and thought text match the principal RL records. These completed files supersede the earlier partial reviews.

The selected cases comprise five robust controls, 12 cases from $FF$ pairs, nine bias-trap ($PF$) cases, and nine cases selected as rescued but still incorrect. These mutually exclusive strata define the failure-enriched sample. Thirteen cases carry a rescue flag across all strata. The automatic evaluator marks ten cases correct and 25 incorrect. 

\begin{table*}[t]
\centering\small
\caption{All completed binary review axes. ``Yes'' follows the original rubric; for thought--answer consistency, Yes affirms consistency. Every count has denominator 35. Agreement compares the clinicians' ratings, not agreement between a clinician and the automatic evaluator. $\kappa$ is computed from the corresponding two-by-two table.}
\label{tab:doctor_full}
\begin{tabular}{@{}lrrrrr@{}}
\toprule
Rubric item & Clinician 1 Yes & Clinician 2 Yes & Both Yes & Agree / 35 & $\kappa$\\
\midrule
Agreement with automatic verdict & 26 & 33 & 26 & 28 & 0.298\\
Overthinking & 15 & 12 & 7 & 22 & 0.222\\
Underthinking & 13 & 3 & 1 & 21 & $-0.017$\\
Missed critical information & 3 & 1 & 0 & 31 & $-0.045$\\
Misweighted finding & 4 & 1 & 0 & 30 & $-0.048$\\
Hallucinated finding & 2 & 0 & 0 & 33 & 0.000\\
Thought--answer consistency & 26 & 34 & 25 & 25 & $-0.054$\\
\bottomrule
\end{tabular}
\end{table*}

Table~\ref{tab:doctor_full} reports all completed binary review axes. The overthinking rubric requests at least three redundant differential passes without new information, with a supporting quotation. Clinicians flag it jointly in seven cases, in either review in 20, and neither review in 15. Their two-by-two counts are seven both Yes, eight clinician-1-only Yes, five clinician-2-only Yes, and 15 both No. Raw agreement is therefore 22/35, and $\kappa=0.222$ \cite{cohen}. Underthinking and overthinking can co-occur in a case; the rubric does not make them exclusive categories.

\begin{table}[h]
\centering\small
\caption{Clinical acceptability ratings, preserving the ordinal five-point scale. Every clinician rates all 35 cases.}
\label{tab:doctor_acceptability}
\begin{tabular}{@{}lrr@{}}
\toprule
Score / summary & Clinician 1 & Clinician 2\\
\midrule
1 & 0 & 0\\
2 & 6 & 0\\
3 & 8 & 6\\
4 & 16 & 20\\
5 & 5 & 9\\
\midrule
Mean & 3.57 & 4.09\\
Median & 4 & 4\\
\bottomrule
\end{tabular}
\end{table}

Table~\ref{tab:doctor_acceptability} gives the acceptability distributions. Ratings agree exactly on 15/35 cases and are within one point on 29/35. Linearly weighted $\kappa$ is 0.203 \cite{weightedkappa}. Binary kappas are reported alongside raw counts because the marginal flag frequencies differ and several findings are rare. 

Clinician 1's file has a blank pair identifier for CASE\_018 and a blank reference diagnosis for CASE\_029. We join by case identifier and obtain pair membership from MamaBench. All original rating cells are retained; the other shared review fields agree between files.

\subsection{Direction of disagreement with automatic verdicts}
Clinician 1 disputes nine of 35 automatic verdicts; all nine are negative verdicts, with no positive verdict disputed. Clinician 2 disputes two negative verdicts and no positive verdicts. The two clinicians jointly dispute CASE\_243 and CASE\_244, the members of PAIR\_122. These disagreements concern negative automatic verdicts in the reviewed sample. Figure~\ref{fig:doctor} summarizes the clinicians' behavioral flags, retaining separate counts for each reviewer.

Of the 25 automatic negative verdicts in the sample, clinician 1 disputes nine and clinician 2 disputes two; both dispute CASE\_243 and CASE\_244. CASE\_029, whose reference field is blank in clinician 1's file, is among clinician 1's nine disagreements. The primary benchmark scores retain the archived labels.

\begin{figure}[t]
\centering
\begin{tikzpicture}
\begin{axis}[paperaxis,width=\linewidth,height=4.8cm,ybar,bar width=13pt,
 ymin=0,ymax=20,ylabel={Flagged cases (of 35)},
 symbolic x coords={Overthinking,Underthinking,Inconsistency},xtick=data,
 x tick label style={font=\scriptsize,rotate=15,anchor=east},ymajorgrids=true,
 nodes near coords,nodes near coords style={font=\scriptsize},enlarge x limits=.25,
 legend style={at={(.5,1.02)},anchor=south,legend columns=2,font=\scriptsize,draw=none}]
\addplot[fill=ink!70,draw=none] coordinates {(Overthinking,15)(Underthinking,13)(Inconsistency,9)};
\addplot[fill=evidence,draw=none] coordinates {(Overthinking,12)(Underthinking,3)(Inconsistency,1)};
\legend{Clinician 1,Clinician 2}
\end{axis}
\end{tikzpicture}
\caption{Clinician flags on the same 35 selected cases. Inconsistency is the inverse of the rubric's thought--answer consistency item; the other bars count affirmative flags.}
\label{fig:doctor}
\end{figure}

\section{Secondary Protocol Comparisons}
\label{app:secondary}
\subsection{Historical direct-thinking failures}
\begin{table*}[t]
\centering\small
\caption{Within-run answer-availability yields. Unanswered outputs lack an accepted prediction at the lower prefix, including earlier-stopped outputs. New answers become available at the upper prefix. No accepted-prefix losses occur at these boundaries.}
\label{tab:interval_yields}
\begin{tabular}{@{}llrrr@{}}
\toprule
Protocol & Token interval & Unanswered at start & New answers & Yield (\%)\\
\midrule
Clarified bounded thinking & 1,000--2,000 & 133 & 31 & 23.3\\
Clarified bounded thinking & 2,000--3,000 & 102 & 8 & 7.8\\
\midrule
Historical direct thinking & 1,000--2,000 & 309 & 79 & 25.6\\
Historical direct thinking & 2,000--3,000 & 230 & 20 & 8.7\\
Historical direct thinking & 3,000--4,096 & 210 & 17 & 8.1\\
\bottomrule
\end{tabular}
\end{table*}
The earlier obstetric-framed direct-thinking run contains 193 extraction failures among 434 cases (44.5\%): 134 JSON parsing failures and 59 blank/template predictions. It emits accepted predictions for both members of only 70/217 pairs and is jointly correct on 37, giving 17.1\% robust accuracy on the full benchmark but 52.9\% conditional accuracy on those 70 available pairs. Its companion direct-answer run scores 50.7\% with all cases available. 

An earlier bounded-thinking run has 187 observed continuations (43.1\%) and 51.6\% robust accuracy. The principal run uses 94 continuations (21.7\%) and reaches 63.6\%. These snapshots record separate prompt and stopping configurations; Table~\ref{tab:interval_yields} gives the within-run prefix yields.

\subsection{Paired diagnostic performance}
Table~\ref{tab:main} reports all 217 pairs for each of the four principal conditions. Each end-to-end condition has 434 case records and no remaining error state after its configured emission procedure. The RL checkpoint under bounded thinking reaches 138 jointly correct pairs, corresponding to 63.6\% robust accuracy. Base accuracy is 82.9\%, whereas counterfactual accuracy is 73.3\%. Thus, performance on isolated base cases substantially exceeds performance requiring correctness on both presentations. The 42 base-correct, counterfactual-incorrect pairs yield a BTR of 23.3\%.

\begin{table*}[t]
\centering\small
\caption{MamaReason on MamaBench: all conditions cover 217 pairs / 434 cases. DA denotes direct answering; BT+R denotes bounded thinking with conditional rescue. Values are percentages; pair counts precede the metrics. Arrows indicate preferred direction. SFT rows retain archived labels except the documented CASE\_085 correction; scoring-snapshot details are given in Appendix~\ref{app:provenance}.}
\label{tab:main}
\begin{tabular}{@{}llrrrrrrrr@{}}
\toprule
Checkpoint & Protocol & $PP$ & $PF$ & $FP$ & $FF$ & Base $\uparrow$ & CF $\uparrow$ & Robust $\uparrow$ & BTR $\downarrow$\\
\midrule
SFT & DA & 113 & 48 & 33 & 23 & 74.2 & 67.3 & 52.1 & 29.8\\
SFT & BT+R & 132 & 41 & 27 & 17 & 79.7 & 73.3 & 60.8 & 23.7\\
RL & DA & 116 & 48 & 31 & 22 & 75.6 & 67.7 & 53.5 & 29.3\\
RL & BT+R & 138 & 42 & 21 & 16 & 82.9 & 73.3 & 63.6 & 23.3\\
\bottomrule
\end{tabular}
\end{table*}

\begin{figure*}[t]
\centering
\begin{minipage}[t]{.48\textwidth}
\centering
\begin{tikzpicture}
\begin{axis}[paperaxis,width=\linewidth,height=5.0cm,ybar,bar width=10pt,
  ymin=0,ymax=100,ylabel={Accuracy (\%)},symbolic x coords={SFT DA,SFT BT+R,RL DA,RL BT+R},
  xtick=data,x tick label style={font=\scriptsize,rotate=18,anchor=east},
  ymajorgrids=true,enlarge x limits=.15,legend columns=2,
  legend style={at={(.5,1.03)},anchor=south,legend columns=2,font=\scriptsize,draw=none}]
\addplot[fill=softgray,draw=none] coordinates {(SFT DA,74.2)(SFT BT+R,79.7)(RL DA,75.6)(RL BT+R,82.9)};
\addplot[fill=evidence,draw=none] coordinates {(SFT DA,52.1)(SFT BT+R,60.8)(RL DA,53.5)(RL BT+R,63.6)};
\legend{Base,Robust}
\end{axis}
\end{tikzpicture}
\end{minipage}\hfill
\begin{minipage}[t]{.48\textwidth}
\centering
\begin{tikzpicture}
\begin{axis}[paperaxis,width=\linewidth,height=5.0cm,ybar stacked,bar width=21pt,
  ymin=0,ymax=217,ylabel={Number of pairs},symbolic x coords={SFT DA,SFT BT+R,RL DA,RL BT+R},
  xtick=data,x tick label style={font=\scriptsize,rotate=18,anchor=east},
  ytick={0,50,100,150,200},ymajorgrids=true,enlarge x limits=.17,
  legend style={at={(.5,1.03)},anchor=south,legend columns=4,font=\scriptsize,draw=none}]
\addplot[fill=evidence,draw=none] coordinates {(SFT DA,113)(SFT BT+R,132)(RL DA,116)(RL BT+R,138)};
\addplot[fill=rehearsal!75,draw=none] coordinates {(SFT DA,48)(SFT BT+R,41)(RL DA,48)(RL BT+R,42)};
\addplot[fill=softgray,draw=none] coordinates {(SFT DA,33)(SFT BT+R,27)(RL DA,31)(RL BT+R,21)};
\addplot[fill=ink!80,draw=none] coordinates {(SFT DA,23)(SFT BT+R,17)(RL DA,22)(RL BT+R,16)};
\legend{PP,PF,FP,FF}
\end{axis}
\end{tikzpicture}
\end{minipage}
\caption{Diagnostic accuracy and pair outcomes across protocols. Left: base and robust accuracy. Right: the full $PP$, $PF$, $FP$, and $FF$ distributions. BT+R denotes bounded thinking with conditional rescue.}
\label{fig:mainresults}
\end{figure*}

The SFT checkpoint exhibits the same descriptive protocol ordering: robust accuracy is 52.1\% for direct answering and 60.8\% for bounded thinking. The corresponding RL values are 53.5\% and 63.6\%. Figure~\ref{fig:mainresults} shows both the base--robust gap and the underlying outcome counts. The bounded-thinking checkpoints have identical counterfactual accuracy, despite different base and robust accuracy. A single accuracy number would conceal these differences in where errors occur.

The bounded-thinking RL--SFT difference is six jointly correct pairs, or 2.8 points. Appendix~\ref{app:provenance} records the scoring-snapshot discrepancies relevant to this comparison.

\subsection{Where the protocol changes pair outcomes}
For the RL checkpoint, bounded thinking increases robust accuracy by 10.1 percentage points over direct answering, with a paired-bootstrap 95\% interval of 4.6--15.7 points. Thirty pairs gain joint correctness and eight lose it (exact McNemar $p=0.00047$). Figure~\ref{fig:transitions} shows that these gains come from several initial states: 14 base-only correct pairs, 13 counterfactual-only correct pairs, and three pairs for which both members were originally wrong. Of the 116 direct-answer $PP$ pairs, 108 remain $PP$.

\begin{figure}[t]
\centering
\begin{tikzpicture}[x=.95cm,y=.75cm]
\foreach \r/\c/\n in {0/0/108,0/1/3,0/2/3,0/3/2,1/0/14,1/1/34,1/2/0,1/3/0,2/0/13,2/1/2,2/2/15,2/3/1,3/0/3,3/1/3,3/2/3,3/3/13}{
  \pgfmathtruncatemacro{\shade}{10+70*\n/108}
  \fill[evidence!\shade] (\c,-\r) rectangle +(1,-1);
  \draw[white,line width=1pt] (\c,-\r) rectangle +(1,-1);
  \node[font=\small,text=ink] at (\c+.5,-\r-.5) {\n};
}
\foreach \i/\lab in {0/PP,1/PF,2/FP,3/FF}{
  \node[font=\small] at (\i+.5,.35) {\lab};
  \node[font=\small,anchor=east] at (-.1,-\i-.5) {\lab};
}
\node[font=\small] at (2,.95) {Bounded thinking + rescue};
\node[font=\small,rotate=90] at (-1.05,-2) {Direct answer};
\end{tikzpicture}
\caption{Paired transitions for the RL checkpoint. Rows are direct-answer outcomes and columns are bounded-thinking outcomes for the same 217 pairs. Joint correctness is gained in 30 pairs and lost in eight. Cell counts describe a protocol change, including a larger budget and conditional continuation.}
\label{fig:transitions}
\end{figure}

Bounded thinking leaves 42 $PF$, 21 $FP$, and 16 $FF$ pairs. The transition matrix shows gains in either presentation as well as losses from previously correct pairs.

This contrast compares the complete protocols in Table~\ref{tab:protocols}: direct answering allows 1,024 output tokens; bounded thinking allows 3,000 first-pass tokens and a conditional 512-token continuation.

\section{Published MamaBench Reference Scores}
\label{app:published}
\subsection{Position relative to published MamaBench systems}
Table~\ref{tab:published} reproduces the eight model/configuration scores reported in the MamaBench paper \cite{mamabench}. The published range is 27.2--65.0\% robust accuracy and 20.3--50.4\% BTR. MamaReason's RL bounded-thinking run occupies the upper part of this reported range: 63.6\% robust accuracy and 23.3\% BTR. Figure~\ref{fig:published} places these values alongside the published reference points.

\begin{table}[t]
\centering\scriptsize
\caption{Published MamaBench reference scores, transcribed from its reported model/configuration comparison \cite{mamabench}. All use 217 pairs. These are external reported scores, not controlled reruns. EA-RAG denotes Evidence-Anchored RAG.}
\label{tab:published}
\begin{tabularx}{\linewidth}{@{}Xrrr@{}}
\toprule
Published system & Base & Robust & BTR\\
\midrule
1. GLM-5 & 54.8 & 27.2 & 50.4\\
2. Kimi K2.5 & 76.0 & 45.6 & 40.0\\
3. GPT-4o + RAG & 77.4 & 46.5 & 39.9\\
4. Nemotron-120B & 72.4 & 47.0 & 35.0\\
5. GPT-4o + EA-RAG & 77.0 & 49.3 & 35.9\\
6. GPT-5.4 & 77.4 & 57.6 & 25.6\\
7. Claude Sonnet 4.6 & 82.0 & 60.8 & 25.8\\
8. Claude Sonnet 4.6 + EA-RAG & 81.6 & 65.0 & 20.3\\
\midrule
MamaReason RL, BT+R (this study) & 82.9 & 63.6 & 23.3\\
\bottomrule
\end{tabularx}
\end{table}

\begin{figure}[t]
\centering
\begin{tikzpicture}
\begin{axis}[paperaxis,width=\linewidth,height=5.4cm,xmin=15,xmax=56,ymin=22,ymax=71,
  xlabel={BTR (\%; lower is better)},ylabel={Robust accuracy (\%)},
  xmajorgrids=true,ymajorgrids=true]
\addplot[only marks,mark=*,mark size=2.5pt,color=softgray] coordinates {(50.4,27.2)(40,45.6)(39.9,46.5)(35,47)(35.9,49.3)(25.6,57.6)(25.8,60.8)(20.3,65)};
\addplot[only marks,mark=diamond*,mark size=4pt,color=evidence] coordinates {(23.3,63.6)};
\node[font=\scriptsize,anchor=south east,text=ink] at (axis cs:50.4,27.2) {1};
\node[font=\scriptsize,anchor=south west,text=ink] at (axis cs:40,45.6) {2};
\node[font=\scriptsize,anchor=north west,text=ink] at (axis cs:39.9,46.5) {3};
\node[font=\scriptsize,anchor=north east,text=ink] at (axis cs:35,47) {4};
\node[font=\scriptsize,anchor=south east,text=ink] at (axis cs:35.9,49.3) {5};
\node[font=\scriptsize,anchor=north west,text=ink] at (axis cs:25.6,57.6) {6};
\node[font=\scriptsize,anchor=south west,text=ink] at (axis cs:25.8,60.8) {7};
\node[font=\scriptsize,anchor=south east,text=ink] at (axis cs:20.3,65) {8};
\node[font=\scriptsize,anchor=north east,text=evidence] at (axis cs:23.3,63.6) {M};
\end{axis}
\end{tikzpicture}
\caption{Published MamaBench systems (gray, indexed as in Table~\ref{tab:published}) and MamaReason RL bounded thinking (teal, M). The upper-left direction favors higher robust accuracy and lower BTR. Systems use different models and inference configurations.}
\label{fig:published}
\end{figure}

The published reference points span different model scales, prompts, retrieval settings, and inference budgets. Their shared pattern is the gap between base-only correctness and success on both presentations.

\section{Conditional Rescue Outcomes}
\subsection{Answer emission and conditional rescue}
The saved RL bounded-thinking records mark 94 of 434 cases as successfully rescued; 54 of these are diagnostically correct (57.4\%). Among the 340 cases without that flag, 285 are correct (83.8\%). At pair level, 82 pairs contain at least one successfully rescued case, and 40 are jointly correct (48.8\%). The other 135 pairs contain 98 jointly correct outcomes (72.6\%). Figure~\ref{fig:rescue} shows these conditional quantities.

\begin{figure}[t]
\centering
\begin{tikzpicture}
\begin{axis}[paperaxis,width=\linewidth,height=4.5cm,ybar,bar width=22pt,
  ymin=0,ymax=100,ylabel={Correct (\%)},
  symbolic x coords={Case rescued,Case unrescued,Pair any rescued,Pair none rescued},xtick=data,
  x tick label style={font=\scriptsize,rotate=22,anchor=east},ymajorgrids=true,
  nodes near coords,nodes near coords style={font=\scriptsize},enlarge x limits=.2]
\addplot[fill=evidence!80,draw=none] coordinates {(Case rescued,57.4)(Case unrescued,83.8)(Pair any rescued,48.8)(Pair none rescued,72.6)};
\end{axis}
\end{tikzpicture}
\caption{RL correctness by rescue status. Case denominators are 94 and 340; pair denominators are 82 and 135. Rescued pairs contain at least one rescued member. Both members must be correct for pair success.}
\label{fig:rescue}
\end{figure}

The rescue flags coincide with the continuation markers and replayed first-pass failures. Section~\ref{sec:emission_results} reports their readouts before continuation.

\section{Trace Examples and Stratified Repetition}
\label{app:traces}
\subsection{Reasoning trace analysis}
Records contain raw generation, parsed prediction, visible rationale, verdict, error state, and rescue status. We check excerpt figures against the original saved text and separate first-pass thought from synthetic channel closure and continuation.

We distinguish three observable behaviors. \emph{Evidence-sensitive deliberation} compares alternatives using case details and develops an evidential argument. \emph{Hypothesis revision} changes an interpretation after reconsidering evidence; beneficial self-correction additionally requires an appropriate emitted answer. \emph{Rehearsal} revisits substantially the same differential or conclusion without a substantive evidential update. These behaviors can coexist. Figure~\ref{fig:rehearsal} presents a clear repetitive suffix, whereas Figure~\ref{fig:motivation} illustrates revision mixed with later oscillation. These definitions concern visible text \cite{faithfulness}.

\begin{figure*}[t]
\centering
\begin{minipage}[t]{0.485\linewidth}
\tracepanel{ink}{A\quad Base presentation\hfill CASE\_365}{%
\textbf{Shared setting:} term newborn with delayed meconium passage.\par\smallskip
Three days old; no meconium since birth; swollen, tense abdomen; feeding poorly.\par\smallskip
\textbf{Benchmark reference:} Hirschsprung's disease.}
\end{minipage}\hfill
\begin{minipage}[t]{0.485\linewidth}
\tracepanel{evidence}{B\quad Counterfactual presentation\hfill CASE\_366}{%
\textbf{Shared setting:} term newborn with delayed meconium passage.\par\smallskip
Two days old; stool after mild interventions; soft, mildly distended abdomen; feeding well.\par\smallskip
\textbf{Benchmark reference:} meconium plug syndrome.}
\end{minipage}
\par\medskip
\tracepanel{evidence}{C\quad Evidence-sensitive revision in the counterfactual thought channel}{%
\textit{``Let's reconsider: could it be meconium plug syndrome?''}\par\smallskip
\textit{``This fits the `stool started after some mild interventions' and `feeding well and generally active' description better.''}}
\par\smallskip
\begin{minipage}[t]{0.60\linewidth}
\tracepanel{rehearsal}{D\quad Later oscillation in the same first pass}{%
\textit{``Let's try Hirschsprung disease.''}\par
\textit{``Let's try meconium plug syndrome.''}\par\smallskip
The trace repeatedly reopens the alternatives before first-pass generation ends.}
\end{minipage}\hfill
\begin{minipage}[t]{0.37\linewidth}
\tracepanel{ink}{E\quad Parsed answer after rescue}{%
\textbf{Hirschsprung disease}\par\smallskip
The continuation emits a diagnosis different from the benchmark reference for CASE\_366.}
\end{minipage}
\caption{PAIR\_183 and selected CASE\_366 trace excerpts. Narratives are abbreviated; thought excerpts are noncontiguous. The counterfactual reference is meconium plug syndrome, whereas the rescue emits Hirschsprung disease.}
\label{fig:motivation}
\end{figure*}
\begin{figure*}[t]
\centering
\begin{minipage}[t]{0.465\linewidth}
\tracepanel{evidence}{A\quad Earlier evidence comparison\hfill CASE\_116}{%
The trace weighs a reported midline defect against the absence of a covering sac and compares gastroschisis with omphalocele.\par\smallskip
It explicitly recognizes the tension between these observations before settling on a candidate diagnosis.}
\end{minipage}\hfill
\begin{minipage}[t]{0.505\linewidth}
\tracepanel{rehearsal}{B\quad Later recurring suffix in the same thought}{%
\textit{``Let's go with gastroschisis.''}\par
\textit{``Wait, let me check if there's any other condition. Maybe gastroschisis? Yes.''}\par
\textit{``Let's go with gastroschisis.''}\par
\textit{``Wait, let me check if there's any other condition. Maybe omphalocele? No.''}\par
\textit{``Let's go with gastroschisis.''}}
\end{minipage}
\par\smallskip
\tracepanel{ink}{C\quad Separate answer-rescue continuation}{%
Parsed diagnosis: \textbf{gastroschisis}. The saved continuation also repeats answer material.}
\caption{Reassertion in CASE\_116. Panel A summarizes earlier evidence comparison; panel B quotes a recurring suffix; panel C gives the separately generated continuation. The suffix repeatedly selects a diagnosis without adding a vignette observation.}
\label{fig:rehearsal}
\end{figure*}
For quantitative description, we lowercase first-pass thought, retain alphabetic words with internal apostrophes, and form the multiset $\mathcal{G}_4$ of consecutive four-word windows. The repeated four-gram fraction is
\begin{equation}
r_4=1-\frac{\lvert\operatorname{unique}(\mathcal{G}_4)\rvert}{\lvert\mathcal{G}_4\rvert}.
\end{equation}
Each bounded-thinking arm contains 434 nonempty traces. Table~\ref{tab:repetition_stratified} stratifies $r_4$ by rescue and final correctness. For RL, rescued correct and incorrect responses have medians of 0.503 and 0.491; unrescued responses have medians of 0.047 and 0.055. Whole-trace $r_4$ varies with length as well as recurrence, so the suffix assertion analysis supplies a complementary view.

\begin{table}[t]
\centering\small
\caption{Whole-thought repeated four-gram fraction by rescue status and final correctness. Continuation text is excluded.}
\label{tab:repetition_stratified}
\begin{tabular}{@{}lllrr@{}}
\toprule
Model & Rescued & Correct & $n$ & Median $r_4$\\
\midrule
SFT & No & No & 62 & 0.046\\
SFT & No & Yes & 279 & 0.046\\
SFT & Yes & No & 40 & 0.474\\
SFT & Yes & Yes & 53 & 0.429\\
RL & No & No & 55 & 0.055\\
RL & No & Yes & 285 & 0.047\\
RL & Yes & No & 40 & 0.491\\
RL & Yes & Yes & 54 & 0.503\\
\bottomrule
\end{tabular}
\end{table}
\section{Evaluation Records and Verdict Consistency}
\label{app:provenance}
The principal analysis uses the saved clarified-prompt SFT and RL direct-answer and bounded-thinking records, each with all 434 cases. The original exports are preserved. A derived analysis copy corrects the SFT direct-answer verdict for CASE\_085 from incorrect to correct, matching the archived final decision for the identical prediction, ``Acute otitis media with perforation.'' This changes one pair from $FP$ to $PP$: SFT direct answering has 113 rather than 112 jointly correct pairs. No other verdict is changed in the reported primary analysis.

Six identical-prediction cases differ in verdict between SFT and RL direct-answer snapshots, and five between the bounded-thinking snapshots. Within RL, CASE\_407 is the sole identical-prediction disagreement between direct answering and bounded thinking; within SFT there are none. A case-label change can alter robust accuracy by at most $100/217=0.46$ points. The first-pass-versus-continuation replay uses one set of bounded-thinking labels throughout.

We recompute pair metrics from case records and check odd/even base--counterfactual membership against MamaBench. Replay reproduces 94 RL and 93 SFT first-pass failures; every marked continuation is accepted. Quantitative readouts use all eligible records, while trace illustrations use selected, verified excerpts.

Replay also reproduces the historical direct-thinking run's 193 error states. Published scores retain the source paper's system labels and remain separate from the local paired resampling. Appendix~\ref{app:training} gives the training settings.

\end{document}